\pdfoutput=1
\documentclass[sigplan,screen]{acmart}

\usepackage{microtype}
\usepackage[T1]{fontenc}
\usepackage[utf8]{inputenc}
\usepackage{newunicodechar}
\newunicodechar{ℍ}{\ensuremath{\mathbb{H}}}
\newunicodechar{μ}{\ensuremath{\mathrm{\mu}}}
\newunicodechar{α}{\ensuremath{\mathrm{\alpha}}}
\newunicodechar{λ}{\ensuremath{\mathrm{\lambda}}}
\newunicodechar{θ}{\ensuremath{\mathrm{\theta}}}
\newunicodechar{ν}{\ensuremath{\mathrm{\nu}}}
\newunicodechar{γ}{\ensuremath{\mathrm{\gamma}}}
\newunicodechar{δ}{\ensuremath{\mathrm{\delta}}}
\newunicodechar{η}{\ensuremath{\mathrm{\eta}}}
\newunicodechar{ε}{\ensuremath{\mathrm{\epsilon}}}
\newunicodechar{≠}{\ensuremath{\neq}}
\newunicodechar{ℝ}{\ensuremath{\mathbb{R}}}
\newunicodechar{∀}{\ensuremath{\forall}}
\newunicodechar{∃}{\ensuremath{\exists}}
\newunicodechar{≤}{\ensuremath{\leq}}
\newunicodechar{≥}{\ensuremath{\geq}}
\newunicodechar{∧}{\ensuremath{\land}}
\newunicodechar{ℕ}{\ensuremath{\mathbb{N}}}
\newunicodechar{→}{{\ensuremath{\rightarrow}}}
\newunicodechar{⊆}{{\ensuremath{\subseteq}}}
\usepackage{xspace}
\usepackage{bbm}
\usepackage{xcolor}
\usepackage{amsmath}

\newcommand{\lean}{{\tt Lean}\xspace}
\newcommand{\mathlib}{{\tt mathlib}\xspace}

\newcommand{\support}{\mathcal{X}}

\newcommand{\concepts}{\mathcal{C}}
  
\newcommand{\target}{t}

\ifdefined\B
\renewcommand{\B}{\mathbb{B}}
\else
\newcommand{\B}{\mathbb{B}}
\fi

\newcommand{\alg}{\mathcal{A}}

\newcommand{\sampled}{\sim}
\newcommand{\classify}{\textsc{Label}}
\newcommand{\lclassify}{\textsc{LList}}
\newcommand{\error}{\textsc{Error}}

\newcommand{\maplist}{\textsc{Map}}

\newcommand{\meas}{{\bf {\tt Meas}}}
\theoremstyle{definition}
  \newtheorem{theorem}{Theorem}[section]
\theoremstyle{definition}
  \newtheorem{counterex}{Counterexample}[section]

\setcopyright{acmlicensed}
\acmPrice{15.00}
\acmDOI{10.1145/3437992.3439917}
\acmYear{2021}
\copyrightyear{2021}
\acmSubmissionID{poplws21cppmain-p15-p}
\acmISBN{978-1-4503-8299-1/21/01}
\acmConference[CPP '21]{Proceedings of the 10th ACM SIGPLAN International Conference on Certified Programs and Proofs}{January 18--19, 2021}{Virtual, Denmark}
\acmBooktitle{Proceedings of the 10th ACM SIGPLAN International Conference on Certified Programs and Proofs (CPP '21), January 18--19, 2021, Virtual, Denmark}


\begin{document}

\propertitle{A Formal Proof of PAC Learnability for Decision Stumps}
\title[A Formal Proof of PAC Learnability for Decision Stumps]{A Formal Proof of PAC Learnability\\ for Decision Stumps\\
  \emph{\small Dedicated to Olivier Danvy on the occasion of his sixtieth birthday.}
}

\author{Joseph Tassarotti}
\email{tassarot@bc.edu}
\affiliation{%
  \institution{Boston College}
  \country{USA}
}

\author{Koundinya Vajjha}
\email{kov5@pitt.edu}
\affiliation{%
  \institution{University of Pittsburgh}
  \country{USA}
}

\author{Anindya Banerjee}
\email{anindya.banerjee@imdea.org}
\affiliation{%
  \institution{IMDEA Software Institute}
  \country{Spain}
}

\author{Jean-Baptiste Tristan}
\email{tristanj@bc.edu}
\authornote{Corresponding author}
\affiliation{%
  \institution{Boston College}
  \country{USA}
}

\begin{abstract}
We present a formal proof in \lean of probably approximately correct (PAC)
learnability of the concept class of decision stumps. This classic result in
machine learning theory derives a bound on error probabilities for a simple type
of classifier.  Though such a proof appears simple on paper, analytic and
measure-theoretic subtleties arise when carrying it out fully formally.  Our
proof is structured so as to separate reasoning about deterministic properties of
a learning function from proofs of measurability and analysis of probabilities.

 \end{abstract}

\maketitle

\section{Introduction}\label{section:introduction}

Machine learning (ML) has achieved remarkable success in a number of problem
domains. However, the often opaque nature of ML has led to concerns about its
use in important contexts such as medical diagnosis and fraud detection.  To address these concerns, researchers have developed a number
of algorithms with proved guarantees of
robustness, privacy, fairness, and accuracy.

One essential property for an ML algorithm is to \emph{generalize} well to
unseen data.  That is, after an algorithm has been trained on some data, it
should be possible to present it with new data and expect it to classify or
analyze it correctly. \citeauthor{Valiant84} introduced the framework of
Probably Approximately Correct (PAC) learnability~\citep{Valiant84}, which gives a mathematical
characterization of what it means for an algorithm to generalize well.  This
framework has become an essential part of the study of computational and
statistical learning theory, and a large body of theoretical results has been
developed for proving that an algorithm generalizes.

However, at present, the vast majority of these and other proofs in ML theory are pencil-and-paper
arguments about idealized versions of algorithms. There is considerable
room for error when real systems are built based on these
algorithms.  Such errors can go unnoticed for long periods of time, and are
often difficult to diagnose with testing, given the randomized behavior of ML
systems. Moreover, the original pencil-and-paper proofs of correctness may
have errors. Because machine learning algorithms often involve randomized sampling
of continuous data, their formal analysis usually requires measure-theoretic reasoning, which
is technically subtle.

Formal verification offers a way to eliminate bugs from analyses of algorithms
and close the gap between theory and implementation. However, the mathematical
subtleties that complicate rigorous pencil-and-paper reasoning about ML
algorithms also pose a serious obstacle to verification.  In particular, while
there has been great progress in recent years in formal proofs about randomized
programs, this work has often been restricted to \emph{discrete} probability
theory. In contrast, machine learning algorithms make heavy use of both
discrete and continuous data, a mixture that requires measure-theoretic
probability.

Thus, to even begin formally verifying ML algorithms in a theorem prover,
results from measure theory must be formalized first.
In recent years, some libraries of
measure-theoretic results have been developed in various theorem provers~\citep{Holzl13, mathlib-cpp, mathcomp-analysis, coq-proba, coq-markov}.  However, it can be challenging to
tell which results needed for ML algorithms are missing from these libraries.
The difficulty is that, on the one hand, standard textbooks on the
theoretical foundations of machine learning~\citep{shalev2014understanding, kearns1994introduction, mohri2018foundations} omit practically all
measure-theoretic details. Meanwhile, research monographs that give completely
rigorous accounts~\citep{dudley_uclt}  present results in maximal
generality, well beyond what appears to be needed for many ML applications. This
generality comes at the cost of mathematical prerequisites that go beyond even a
first or second course in measure theory.

This paper describes the formal verification in \lean~\cite{MouraKADR15} of a standard result of
theoretical machine learning which illustrates the complexities of measure-theoretic probability. We consider a simple type of classifier called a \emph{decision stump}. A decision stump
classifies real-numbered values into two groups with a simple rule: all
values $\leq$ some threshold $\target$ are labeled $1$, and those above
$\target$ are labeled $0$. We show that decision stumps are PAC learnable by proving a \emph{generalization} bound, which bounds
the number of training examples needed to obtain a chosen level of classification accuracy with high probability.

We describe more precisely below how decision stumps are trained and the bound
that we have proved (\autoref{sec:decision-stumps}). Although decision stumps appear simple, they are worth
considering because they are the 1-dimensional version of a classifying
algorithm for axis-aligned rectangles that is used as a motivating example in all of the standard textbooks
in this
field~\citep{kearns1994introduction,shalev2014understanding,mohri2018foundations}.

In spite of the seeming simplicity of this example, all three of the cited
textbooks either give an incorrect proof of this result or omit
what we found to be the most technically challenging part of the proof (\autoref{sec:informal}). In noting this, we
do not wish to exaggerate the importance of these errors. The proofs can be
fixed, and in each book, the results are correctly re-proven later as consequences of
general theorems. Rather our point is to
emphasize that even basic results in this area can touch on subtle issues, and
errors can evade notice despite much review and scrutiny by a wide
audience. We believe this further motivates the need for machine-checked proofs
of such results if they are to be deployed in high-importance settings.

A key component of our work is that we structure our formal proof in a manner
that lets us separate the high-level reasoning found in textbook
descriptions from the low-level details about measurability. We outline the structure used to achieve this separation of concerns in \autoref{sec:structure}. We then describe some preliminaries about measure-theoretic probability in \autoref{sec:prelim}.
The Giry monad~\citep{giry1982categorical} allows us to give a precise description of
the sources of randomness involved in training and evaluating the performance of
classifiers~(\autoref{sec:giry}). We exploit this to split deterministic reasoning about basic
properties of the stump learning algorithm~(\autoref{sec:pure}) from proofs of measurability~(\autoref{sec:measurability}) and the analysis of bounds on probabilities~(\autoref{sec:probabilistic}).

The proof is publicly available at \url{https://github.com/jtristan/stump-learnable}.

\section{Decision Stumps and PAC Learnability}
\label{sec:decision-stumps}
To motivate decision stumps and the result that we have formalized, consider the
following scenario. Suppose a scientist has developed a test to measure levels
of some protein in blood in order to diagnose a disease. 
Assume there is some (unknown) threshold $\target \in \mathbb{R}$ such that if
the protein level is $\leq \target$, then the patient has the disease, and
otherwise does not.  Given a random sample of patients whose disease status is
known, the scientist wants to estimate the threshold $\target$ so that the test can be used to screen and
diagnose future patients whose disease statuses are unknown.

In other words, the scientist wants to find a decision stump to
classify whether patients have the disease or not.  We can model the blood test
as returning a nonnegative real number. There is some distribution $\mu$ on the
interval $[0,\infty)$ representing levels of the protein in the population. The
scientist has samples $x_1, \dots, x_n \in [0, \infty)$ independently drawn from
$\mu$ giving the results of the test on a collection of $n$ patients, along with
labels $y_1, \dots, y_n \in \{0, 1\}$ giving each patient's disease status.
A label of $1$ means a patient has the disease, while $0$ means they do
not, so that $y_i = 1$ if and only if $x_i \leq \target$.  The $(x_1,
y_1), \dots, (x_n, y_n)$ are called \emph{training examples}.  The scientist is
trying to pick some value $\hat{\target}$ as an estimate of $\target$ to use to
classify future patients. In particular, she will use her estimate to define a decision stump classifier.
To state this formally, we first define a function $\classify$ that
assigns a label to a point $x$ according to a threshold $d$:
\begin{align}
\classify(d, x) =
\begin{cases}
1 &\quad \text{if $x \leq d$} \\
0 &\quad \text{if $x > d$}
\end{cases}
\end{align}
The scientist will pick some threshold $\hat{t}$ and then use the classifier $\lambda x.\, \classify(\hat{t}, x)$ to label future patients. We call such a classifier a \emph{hypothesis}.

How should the scientist select $\hat{\target}$? One idea is to take $\hat{\target}$ to be
the maximum of the $x_i$ that have label $1$. (If no $x_i$ has label $1$, she can take $\hat{\target}$ to be $0$.) This estimate, at least, would
correctly label all of the training examples. This corresponds to the following \emph{learning algorithm} $\alg$, which returns a classifier using this estimate:
\newcommand{\bind}[3]{\textsf{let}\;#1\;\textsf{=}\,#2\;\textsf{in}\;#3}
\begin{align}
\begin{split}
&\alg\left([(x_1, y_1),\dots,(x_n,y_n)]\right)\\
&=
\bind{\hat{t}}{\max\{x_i \ |\ y_i = 1\}}{} \\
&\quad\lambda x.\, \classify(\hat{t}, x)
\end{split}
\end{align}
where $\max$ of the empty set is defined to be $0$.\footnote{We call this function an \emph{algorithm} here, following \citet{shalev2014understanding}, although the operations on real numbers involved are non-computable.}
Of course, the estimate $\hat{\target}$ used in the classifier
returned by this algorithm is not going to be exactly the value of $\target$, especially if the number of
training examples, $n$, is small. But if $n$ is large enough, we might hope that
a good estimate can be produced. The key question then becomes, how many
training examples should the scientist use?

To answer this more precisely, we need to decide how to evaluate the quality of
the classifier returned by $\alg$. At first, one might think that the goal should be
to minimize $|\hat{\target} - \target|$, that is, to get an estimate that is as
close as possible to the true threshold $\target$. While minimizing the distance
between $\hat{\target}$ and $\target$ is useful, our primary concern
should be how well we classify future examples.
The $\error$ of a classifier $h$ is the probability that $h$ mislabels
a test example $x$ randomly sampled from $\mu$:
\begin{equation}
\label{eqn:error-def}
\error(h) = \Pr_{x \sampled \mu}\left(h(x) \neq \classify(\target, x)\right)
\end{equation}
We write %
$x \sampled \mu$ in the above to %
indicate that the random variable $x$ has distribution $\mu$.
This $x$ is independent of the training examples used by the scientist.
While the definition of $\error$ refers to this randomized scenario of drawing
a test sample, for a fixed hypothesis $h$ the quantity $\error(h)$ is a real number.

Because the training examples the scientist uses are randomly selected, the $\error$ of the hypothesis
she selects using $\alg$ is a random variable.
In practice, the scientist does not know either the distribution $\mu$ or the target $\target$, so she cannot
compute the exact $\error$ of the classifier she obtains.
Nevertheless, she might want to try to ensure that the $\error$ of the
classifier she defines using $\alg$ will be below some small $\epsilon$ with high probability.

To talk about the $\error$ of the selected hypothesis precisely,
let us first introduce a helper function which takes an unlabeled list of examples and returns a list where each example has been paired with its true label:
\begin{align}
\begin{split}
&\lclassify([x_1, \dots, x_n]) \\ &\quad= \maplist\ \bigl(\lambda x.\, (x, \classify(t, x))\bigr)\ [x_1, \dots, x_n]
\end{split}
\end{align}
Then through her choice of $n$, the scientist can bound the following probability:  
\[ \Pr_{(x_1, \dots, x_n) \sampled \mu^n}\bigl(\error(\alg(\lclassify([x_1, \dots,x_n]))) \leq \epsilon\bigr) \]
where $(x_1, \dots, x_n) \sampled \mu^n$ indicates that the variables $x_i$ are drawn independently
from the same distribution $\mu$.
The following theorem, which is the central result that we have formalized,
tells the scientist how to select $n$ to achieve a desired bound on this probability:

\begin{theorem}
\label{thm:informal}
  For all $\epsilon$ and $\delta$ in the open interval $(0, 1)$, if $n \geq \frac{\ln(\delta)}{\ln(1 - \epsilon)} - 1$ then
  \begin{equation}
  \label{eqn:informal-bound}
 \Pr_{(x_1, \dots, x_n) \sampled \mu^n}\bigl(\error(\alg(\lclassify([x_1, \dots,x_n]))) \leq \epsilon\bigr)
   \geq 1 - \delta
  \end{equation}
\end{theorem}

Before giving an informal sketch of how this theorem is proved, we briefly
describe how this result fits into the %
framework of PAC learnability~\citep{Valiant84}.

\paragraph{PAC Learnability.}

PAC learning theory gives an abstract way to explain  scenarios like the one
with the scientist described above. In this general set up, there is some set
$\support$ of possible examples and a set $\concepts$ of hypotheses $h : \support \rightarrow \{0, 1\}$,
which are possible classifiers we might select. The set $\concepts$ is %
also called a \emph{concept class}. %
In the case of decision stumps,
$\support = \mathbb{R}^{\geq 0}$, and $\concepts = \{ \lambda x.\, \classify(d, x) \mid d \in \mathbb{R}^{\geq{0}}\}$.

As in the stump example, there is assumed to be some unknown distribution $\mu$
over $\support$.  Additionally, there is some function $f : \support \to \{0,
1\}$ that maps examples to their true labels.
The $\error$ of a hypothesis
is the probability that it incorrectly labels an example drawn according to $\mu$. 
The function $f$ is said to
be \emph{realizable} if there is some hypothesis $h \in \concepts$ that has error $0$.
The goal is to select a hypothesis with minimal $\error$ when given a collection
of training examples that have been labeled according to $f$. A concept class is
said to be PAC learnable if there is some algorithm to select a hypothesis for which
we can compute the number of training examples needed to achieve error bounds with high probability
as in the stump example:

\begin{definition}
\label{def:pac}
A concept class $\concepts$ is PAC learnable if there exists an algorithm $\alg : \textsf{List} (\support \times \{0, 1\}) \rightarrow \concepts$ and a function
$g: (0,1)^2 \to \mathbb{N}$ such that for all distributions $\mu$ on $\support$ and realizable label functions $f$,
when $\alg$ is run on a list of at least $g(\epsilon, \delta)$ independently sampled examples from $\mu$ with labels computed by $f$,
the returned hypothesis $h$ has error $\leq \epsilon$ with probability $\geq 1 - \delta$. 
\end{definition}
By ``there exists'' in the above definition, one typically conveys %
a constructive sense:  one can 
exhibit the algorithm and compute $g$.\footnote{Some authors require further that the algorithm $\alg$
has polynomial running time, but we will not do so.} The function $g$
is a bound on the \emph{sample complexity} of the algorithm, telling us
how many training examples are needed to achieve a bound on the \error\ with a given probability.
There is a large body of theoretical results for showing that a concept class is
PAC learnable and for bounding the sample complexity by analyzing %
the VC-dimension~\citep{vapnik2000} of the concept class.

\autoref{thm:informal} states that the concept class of decision stumps is PAC learnable.
We next turn to how this theorem is proved.

\section{Informal Proof of PAC Learnability}
\label{sec:informal}

The PAC learnability of decision stumps follows from the
general VC-dimension theory alluded to above. However, mechanizing that
underlying theory in all its generality is beyond the scope of this paper.

Instead this paper formalizes a more elementary proof, based
on a description from three
textbooks~\citep{kearns1994introduction, shalev2014understanding,
mohri2018foundations}. The proofs %
in the textbooks are for a slightly different
result---learning a 2-dimensional rectangle instead of a decision stump---but
the idea is essentially the same. We first sketch how the proof is presented
in two of the textbooks~\citep{kearns1994introduction, shalev2014understanding}. As we shall see, this argument has a flaw.

\begin{proof}[Proof sketch of \autoref{thm:informal}]
Given $\mu$ and $\epsilon$, consider labeled samples $(x_1,y_1),...,(x_n,y_n)$.
Recall that the true threshold is some unknown value $t$, and the learning algorithm $\alg$ here returns a classifier that uses the maximum of the
positively labeled examples as the decision threshold $\hat{t}$.

We start by noting some deterministic
properties about this classifier. %
Observe that
$\alg$ ensures $\hat{t} \leq t$, because all positively labeled training
examples must be $\leq t$. Furthermore, a test example $x$ is misclassified
only if $\hat{t} < x \leq t$.  Thus the only errors that the classifier selected
by $\alg$ can make is by incorrectly assigning the label $0$ to an example that should have label $1$.

With this in mind, %
the proof goes by cases on the probability that a randomly sampled
example $x \sampled \mu$ will have true label $1$. First assume
$ \Pr_{x \sampled \mu}\left(x \leq \target\right) \leq \epsilon$. That is, this case
assumes that examples with true label 1 are rare. Then the
classifier returned by $\alg$ must have $\error$ that is $\leq \epsilon$. In other words,
if the returned classifier can only misclassify examples whose true label is $1$, and those are
sufficiently rare, i.e., have probability $\leq \epsilon$ by the above assumption, then the classifier has the desired error bound.

Next assume 
$\Pr_{x \sampled \mu}\left(x \leq \target\right) > \epsilon$.
The idea for this case is to find an
interval $\mathcal{I}$ such that, so long as at least one of the training examples $x_i$ falls
into $\mathcal{I}$, the classifier returned by $\alg$ will have error
$\leq \epsilon$. Then we find a bound on the probability that \emph{none} of the
$x_i$ fall into $\mathcal{I}$.

In particular, set $\mathcal{I} = [\theta,\target]$, choosing $\theta$ so that
\[ \Pr_{x \sampled\mu}\left(x \in \mathcal{I}\right) = \epsilon \] That is, we
want $\mathcal{I}$ to enclose exactly probability $\epsilon$ under $\mu$.  Let
$E$ be the event that at least one of the training examples falls in
$\mathcal{I}$. If $E$ occurs, then the threshold $\hat{t}$ selected by $\alg$ is
in $\mathcal{I}$. To see this, observe that if for some $x_i$ we have $\theta \leq
x_i \leq \target$, then we know $y_i = 1$, and hence $\theta \leq x_i \leq \hat{t} \leq t$.

In that case, for a test example $x$ to be misclassified, we must have $\hat{t} < x \leq t$, meaning $x$ must also lie in $\mathcal{I}$.
Thus, the event of misclassifying $x$ is a subset of the event that $x$ lies in $\mathcal{I}$. Hence, if $E$ occurs, the probability of misclassifying $x$
is at most the probability that $x$ lies in $\mathcal{I}$.
But the probability that $x$ lies in $\mathcal{I}$ is $\epsilon$ by the way we defined $\theta$.
Therefore if $E$ occurs, the probability that a randomly selected example $x$ will be misclassified is $\leq \epsilon$, meaning the $\error$ of the classifier will be $\leq \epsilon$.

So for the error to be above $\epsilon$
means that \emph{none} of our training examples $x_i$ came from $\mathcal{I}$. For each $i$, we have
\[ \Pr_{(x_1, \dots,x_n)}(x_i \notin \mathcal{I}) = 1 - \epsilon \]
Because each $x_i$ is sampled independently from $\mu$, the probability that none of the $x_i$ lie in $\mathcal{I}$ is $(1 - \epsilon)^n$. Thus the probability of $E$, the event that \emph{at least} one $x_i$ is in $\mathcal{I}$, is $1 - (1 - \epsilon)^n$.
Since we have shown that if $E$ occurs, then the $\error$ is $\leq \epsilon$, this means
that the probability that the $\error$ is $\leq \epsilon$ is \emph{at least} the probability of $E$.
The rest of the proof follows by choosing $n$ to ensure $1 - (1 - \epsilon)^n \geq 1 - \delta$.
\end{proof}

The careful reader may notice that there is one subtle step in the above: how do
we choose $\theta$ to ensure that ``$\mathcal{I}$ encloses exactly probability
$\epsilon$ under $\mu$''? The phrasing ``encloses exactly'' comes
from \citet{kearns1994introduction} (page 4), which does not say how to prove that
$\theta$ exists, beyond giving some geometric intuition in which we visualize
shifting the left edge of $\mathcal{I}$ until the enclosed amount has the specified
probability. \citet{shalev2014understanding} similarly instructs us to select
$\theta$ so that the probability ``is exactly'' $\epsilon$.\footnote{The cited
references address the more general problem of axis-aligned rectangles instead
of stumps, so more specifically they describe shifting the edge of a rectangle
until the enclosed probability is $\epsilon/4$.}

Unfortunately, the argument is not correct, because such a $\theta$ may not
exist.\footnote{We are not the first
to observe this error. The errata for the first printing of \citet{mohri2018foundations} points
out the issue in the proof of \citet{kearns1994introduction}.} The following counterexample demonstrates this.

\begin{counterex}
\label{counterex:theta}
Take $\mu$ to be the Bernoulli distribution which returns $1$ with probability
$.5$ and $0$ otherwise. Let $\target = .5$, and $\epsilon = .25$. Then for all $a$ we have:
\begin{align}
\Pr_{x \sampled \mu}(x \in [a, \target]) =
\begin{cases}
.5 \quad \text{if $a \leq 0$} \\
0 \quad \text{otherwise}
\end{cases}
\end{align}
so this means that no matter how we select $\theta$, we cannot have $\Pr_{x \sampled \mu}(x \in [\theta, t]) = \epsilon$, so the desired $\theta$ does not exist.
\end{counterex}
The issue in the proof is that the distribution function $\mu$ has been assumed to be \emph{continuous},
whereas in the above counterexample $\mu$ is a \emph{discrete} distribution. In particular, if there is some point $y$ such that $\Pr_{x \sampled \mu}(x = y) > 0$ then this introduces a \emph{jump discontinuity} in the distribution function.

However, the statement of PAC learnability says that the error bound should be achievable for \emph{any} distribution $\mu$. In order to fix the proof to work for any $\mu$, we need to consider
the following revised definition of
$\theta$, which will ensure it exists:
\begin{equation}
\label{eqn:thetadef}
  \theta = \sup \left\{ d \in \support \mid \Pr_{x \sampled \mu}\left(x \in [d,\target]\right) \geq \epsilon\right\}
\end{equation}
In this definition, the set (say $S$) over which we are taking the supremum might be infinite. However, recall that we only need to construct the point $\theta$ in the sub-case of the proof where we assume that $\Pr_{x \sampled \mu}\left(x \leq \target\right) > \epsilon$. This assumption implies that the supremum exists, because it means that %
$S$ is nonempty, and furthermore we know that $S$ is bounded above by $\target$. The existence of the supremum then follows from the fact that the real numbers are Dedekind complete.

The idea behind this definition of $\theta$ is that, if the distribution function is continuous, then the definition picks a $\theta$ that has the property required in the erroneous proof. Instead if there is a discontinuity that causes a jump in the distribution function past the value $\epsilon$, then the definition selects the point at that discontinuity.
In particular, we can show that with this definition
\begin{equation}
\label{eqn:thetalb}
\Pr_{x \sampled \mu}\left(x \in [\theta,\target]\right) \geq \epsilon
\end{equation}
and \emph{also} that
\begin{equation}
\label{eqn:thetaub}
\Pr_{x \sampled \mu}\left(x \in (\theta,\target]\right) \leq \epsilon
\end{equation}

Note in \autoref{eqn:thetalb} we have the closed interval $[\theta, \target]$,
while \autoref{eqn:thetaub} is about the half-open interval
$(\theta, \target]$. This means that if $\Pr_{x \sampled \mu}(x = \theta) = 0$,
as we would have in a continuous probability distribution, then
$\Pr_{x \sampled \mu}(x \in [\theta, \target]) = \epsilon$. Whereas if $\Pr_{x \sampled \mu}(x = \theta) \neq 0$, as can occur
in a discrete distribution, the probabilities of lying in $[\theta, \target]$ and $(\theta, \target]$ will differ. For example, for the
discrete distribution in \autoref{counterex:theta}, the definition of $\theta$ in \autoref{eqn:thetadef} would yield $\theta = 0$. Observe that $\Pr_{x\sampled \mu}(x=0) = .5 \neq 0$ while $\Pr_{x\sampled \mu}(x=v) = 0$ for any $v\in(0, 1)$.

The original proof of PAC learnability of the class of rectangles~\citep{blumer1989learnability} did give a correct definition of $\theta$, as does the
textbook by \citet{mohri2018foundations}, although neither gives
a proof for why the point defined this way has the desired properties.
Indeed, \citeauthor{mohri2018foundations} say that it is ``not hard to see'' that these properties
hold.

In fact, this turned out to be the most difficult part of the whole proof to
formalize. While it only requires some basic results in measure theory and
topology, it is nevertheless the most technical step of the argument.  There
were two other parts of the proof that seemed obvious on paper but turned out to
be much more technically challenging than expected, having to do with showing
that various functions are measurable.  Often, details about measurability are
elided in pencil-and-paper proofs. This is understandable because these
measurability concerns can be tedious and trivial, and checking that
everything is measurable can clutter an otherwise insightful proof. However,
many important results in statistical learning theory do not hold without
certain measurability assumptions, as discussed
by \citet{blumer1989learnability} and \citet[chapter 5]{dudley_uclt}.

Now that we have seen some intuition for this result and some of the pitfalls in proving it, we describe
the structure of our formal proof and how it addresses these challenges.

\section{Structure of Formal Proof}
\label{sec:structure}

When we examine the informal proof sketched above, we can see that there are
several distinct aspects of reasoning. Instead of intermingling these reasoning steps as in the proof sketch, we structure our formal proof to separate these components. As we will see, this decomposition
is enabled by features of \lean. We believe that this proof structure applies
more generally to other proofs of PAC learnability and related results in ML theory.

Specifically, we identify the following four components. We describe each briefly in the paragraphs below. The remaining sections of the paper then elaborate on each of these parts of the formal proof, after giving
some background on measure theory in \autoref{sec:prelim}.

\paragraph{Specifying sources of randomness:} As
  we saw in \autoref{sec:decision-stumps}, there are two randomized scenarios under
  consideration in the statement of \autoref{thm:informal}. First, there is
  the randomized choice of the training examples that are given as input to
  $\alg$. These are sampled independently and from the
  same distribution $\mu$. This first source of randomness is explicit since it
  appears in the statement of the probability appearing in
  \autoref{eqn:informal-bound}.  The second kind of randomization is in the
  definition of $\error$. Recall that we defined $\error$ in \autoref{eqn:error-def}  as
  the probability that an example randomly sampled from $\mu$ would be
  misclassified. This random sampling is entirely separate from the sampling of
  the training examples, although both samplings utilize the same $\mu$.

  Finally, the theorem statement is quantifying over the distribution
  $\mu$. As we saw in the counterexample to the informal proof, inadvertently
  considering only certain classes of distributions (such as continuous ones)
  leads to erroneous arguments.

  All of these details must be represented formally in the theorem prover. To
  handle these issues, we make use of the Giry monad~\citep{giry1982categorical} which
  allows us to represent sampling from distributions as monadic
  computations. \autoref{sec:giry} explains how this provides a convenient way to
  model the training and testing of a learning algorithm in order to formally
  state \autoref{thm:informal}.

  \paragraph{Deterministic properties of the algorithm:}

  In the beginning of the proof sketch of \autoref{thm:informal} we started by noting
  certain \emph{deterministic} properties of the learning algorithm $\alg$, such
  as the fact that the threshold value $\hat{t}$ in the classifier returned by
  $\alg$ must be $\leq$ the true unknown threshold $t$. These deterministic
  properties were the only details about $\alg$ upon which we relied
  when later establishing the $\error$ bound. This means that an analogue
  of \autoref{thm:informal} will hold for any other stump learning algorithm with those
  properties.

  As we will see, the Giry monad enables us to encode $\alg$ as a purely
  functional \lean term that selects the maximum of the positively labeled
  training examples. This means we can prove these preliminary deterministic
  properties in the usual way one reasons about pure functions in \lean. The \lean statements of these
  deterministic properties are described in \autoref{sec:pure}.

\paragraph{Measurability of maps and events:}

One detail missing from the informal proof was any consideration of
\emph{measurability} of functions and events. In measure-theoretic probability,
probability spaces are equipped with a collection of \emph{measurable sets}. We
can only speak of the probability of an event if we show that the set
corresponding to the event is \emph{measurable}, meaning that it belongs
to this collection. Similarly, random variables, such as the learning algorithm $\alg$
itself, must be \emph{measurable functions}.

While these facts are necessary for a rigorous proof, %
they risk cluttering a formal proof and obscuring all of the intuition
that the informal proof gave. However, with \lean's typeclass mechanism and other
proof automation, we can mostly separate the parts of the proof
concerning measurability from the rest of the argument, as we describe
in \autoref{sec:measurability}.

\paragraph{Quantitative reasoning about probabilities:}

The last step of the proof involves constructing the point $\theta$ described
above and showing bounds on the probability that a sampled example lies in the
interval $[\theta, t]$. Other than correcting the issue involved in the
definition of $\theta$, this stage of reasoning is similar to the proof style found
in informal accounts of this result. Our goal is that this portion of the proof
should resemble the kind of probabilistic reasoning that is familiar to experts
in ML theory. This portion of the argument, and the final proof of \autoref{thm:informal}
are described in \autoref{sec:probabilistic}.

\section{Preliminaries}
\label{sec:prelim}
In this section we describe some basic background on measure-theoretic probability
and how measure theory has been formalized in \lean as part of the \mathlib library~\citep{mathlib-cpp}.

\paragraph{Measure theory.} The starting point for probability theory is a set $\Omega$ called
a \emph{sample space}.  Elements of $\Omega$ are called outcomes, and represent
possible results of some randomized situation. For example, if the randomized
situation is the roll of a six-sided die, we would have $\Omega = \{1, 2, 3, 4,
5, 6\}$.  In naive probability theory, subsets of $\Omega$ are called events,
and a probability function $P$ on $\Omega$ is a function mapping events to real
numbers in the interval $[0, 1]$, satisfying some axioms. While this naive
approach works so long as $\Omega$ is a finite or countable set, attempting to
assign probabilities to \emph{all} subsets of $\Omega$ runs into technical
difficulties when $\Omega$ is an uncountable set such as $\mathbb{R}$.

 Measure-theoretic probability theory resolves this issue by only assigning
probabilities to a collection $\mathcal{F}$ of subsets of $\Omega$. The elements of
$\mathcal{F}$ are called \emph{measurable sets}. This collection $\mathcal{F}$ must
be a \emph{sigma-algebra}, which means that it must be closed under certain
operations (e.g. taking countable unions). We call the pair
$(\Omega, \mathcal{F})$ a \emph{measurable space}. A probability measure $\mu$ is
then a function of type $\mathcal{F} \rightarrow [0,1]$ satisfying the following axioms:

\begin{itemize}

\item $\mu(\emptyset) = 0$
\item $\mu(\Omega) = 1$
\item If $A_1, A_2, \dots$ is a countable collection of measurable sets such that $A_i \cap A_j = \emptyset$ for
$i \neq j$, then
\[ \mu\left(\bigcup_{i=1}^\infty A_i\right) = \sum_{i=1}^\infty \mu(A_i) \]

\end{itemize}

For the reader familiar with topology, the notion of a measurable space is
analogous to the situation in topology, where a topological space is a pair
$(X, \mathcal{V})$ where $X$ is a set and $\mathcal{V}$ is a collection of
subsets of $X$ called the open sets, and $\mathcal{V}$ must be closed under
various set operations.  Indeed, for every topological space there is a minimal
sigma-algebra containing all open sets, which is called the \emph{Borel
sigma-algebra}. We use the Borel sigma-algebra on the real numbers throughout the following.

A function $f: (\Omega_1, \mathcal{F}_1) \rightarrow (\Omega_2, \mathcal{F}_2)$
between two measurable spaces is said to be a \emph{measurable function} if, for
all $A \in \mathcal{F}_2$, we have $f^{-1}(A) \in \mathcal{F}_1$. Again, there
is an analogy to topology, where a function between topological spaces is
continuous if inverse images of open sets are open. In fact, if $f$ is a
continuous function between two topological spaces, then $f$ is measurable when
those spaces are equipped with their Borel sigma-algebras. Continuity implies
that many standard arithmetic operations on the reals are measurable. Other
examples which are measurable but not continuous include functions for testing
whether a real number is $=$, $\leq$ or $\geq$ to some value.

The general measure theory in \mathlib allows the
measures of events to be greater than $1$. To obtain just probability measures,
we restrict these general definitions to require that if $\mu$ is a probability measure on
$X$, then $\mu(X) = 1$. In \autoref{sec:decision-stumps} we subscripted the $\Pr$ notation to %
indicate
the distributions that we were considering. In the context of a theorem prover,
which obliges us to be precise in this manner, we forgo the $\Pr$ notation altogether.
Instead, for the probability of an event $E$ with respect to some measure $\mu$, we simply write $\mu(E)$.

Above we have used the traditional mathematical notation
of writing $f(x)$ for the application of a function $f$ to an argument $x$. However,
to more closely match notation in \lean, %
the sequel uses $f\, x$ when referring
to definitions from our \lean development.

\paragraph{Typeclasses.} In %
mathematical writing, we often associate a particular
mathematical structure, such as a topology or sigma-algebra with a given set,
with the convention that the structure should be used throughout. For example,
when talking about continuous functions from $\mathbb{R} \to \mathbb{R}$,
we do not constantly clarify that we mean continuous functions with respect
to the topology generated by the Euclidean metric on $\mathbb{R}$.

This style of mathematical writing can be mimicked with \lean's typeclasses.
After defining a typeclass, the user can declare instances of that typeclass,
which associate a default structure with a given type.
This mechanism is used throughout
\mathlib to supply default topologies, ring structures, and so on with
particular types. 

For example, the commands below first introduce the notation \texttt{ℍ} to refer to the type \texttt{nnreal} of nonnegative real numbers from \mathlib. We will use this notation
when referring to the type of training examples and thresholds used for classification. After declaring this notation, an instance of \texttt{measurable\_space}
is defined on this type:
\begin{verbatim}
notation `ℍ` := nnreal
instance meas_ℍ: measurable_space ℍ := ...
\end{verbatim}
where we have omitted the definition after the \texttt{:=} sign. After this instance
is declared, any time we refer to \texttt{ℍ} in a context where we need a sigma-algebra, this
instance will be used.

The \mathlib library comes with lemmas to automatically derive instances of {\texttt{measurable\_space}} from other instances.
For example, if a type has been associated with a topology, we can automatically derive the Borel sigma-algebra as an instance of {\tt measurable\_space} for that type. We use this Borel sigma-algebra
on \texttt{ℍ} above. Similarly, we can derive a product sigma-algebra on the product \texttt{A × B} of two types from
existing instances for \texttt{A} and \texttt{B}, as in the following example:
\begin{verbatim}
instance meas_lbl: measurable_space (ℍ × bool) 
\end{verbatim}

\paragraph{Measurable spaces for stump training.}
At this point, considerations about the sigma-algebras with which a type is equipped 
introduce the first discrepancy between the informal set-up
in \autoref{sec:decision-stumps} and our formalization. As we described there,
it is common to treat the learning algorithm as if it returned a \emph{function}
of type $\texttt{ℍ} \to \{0, 1\}$, mapping examples to labels. Since one wants
to speak about probabilities involving these classifiers, this means the type of
classifiers must be equipped with a sigma-algebra. What sigma-algebra should be
chosen? While there are canonical choices for types that have a topology (the
Borel sigma-algebra) and for various operations on spaces such as products,
there is no such standard choice for function types. In particular, the category
of measurable spaces is not Cartesian closed~\citep{Aumann}. Hence, there is no
generic sigma-algebra on function types that would also make evaluation
measurable. The textbook by \citeauthor{shalev2014understanding} points
out that the PAC learnability framework requires the existence of a sigma-algebra on the class of hypotheses
that makes classification measurable~\citep[Remark 3.1]{shalev2014understanding},
but the \emph{construction} of this sigma-algebra is not typically explained in examples.

Fortunately, the subset of decision stump classifiers has a simpler structure
than the type of \emph{all} functions from $\texttt{ℍ} \to \{0, 1\}$. In
particular, the behavior of a decision stump classifier is entirely determined
by the threshold used as a cut-off when assigning labels. These thresholds have
type $\texttt{ℍ}$, which is equipped with the Borel sigma-algebra. In
particular, given a threshold $t$, the function $\lambda x.\, \classify(t, x)$ is
measurable, and this is the evaluation function for a decision stump classifier.
Thus, as we will see in
the next section, we formalize the learning algorithm $\alg$ such that it directly returns a threshold instead of a classifier. Similarly we adjust the definitions of $\error$ (and associated functions) to take a threshold as input instead of a classifier. %

A similar concern arises with how we represent the collection of training
examples passed to the learning algorithm $\alg$. In the earlier informal presentation, %
$\alg$ takes a list of labeled examples as input. However, %
the construction of a sigma-algebra on variable-length lists is not commonly
discussed in measure theory texts. %
We therefore work with dependently
typed vectors of a specified length.  Given a type {\tt A} and natural {\tt n},
the type {\tt vec A n} represents vectors of size {\tt n+1} of values of type
{\tt A}. When $\texttt{A}$ is a topological space, {\tt vec A n} can be given
the ${\tt (n+1)}$-ary product topology, and we can then make it into a measurable space by
equipping it with the Borel sigma-algebra.

\section{Specifying Randomized Processes with the Giry Monad}
\label{sec:giry}

Now that we have described some of the preliminaries of measure-theoretic
probability, we turn to the question of how to formally represent the learning
algorithm in the theorem prover.

In traditional probability theory, it is common to fix some sample space
$\Omega$ and then work with a collection of \emph{random variables} on this
sample space. If $V$ is a measurable space, a $V$-valued random variable is a
measurable function of type $\Omega \to V$. One can think of the elements of the
sample space $\Omega$ as some underlying source of randomness, and then the
random variables encode how that randomness is transformed into an observable
value. For example, $\Omega$ could be a sequence of random coin flips, and a
random variable $f$ might be a randomized algorithm that uses those coin flips.

In fact, at a certain point most treatments of probability theory start to leave
the sample space $\Omega$ completely abstract. One simply postulates the
existence of some $\Omega$ on which a collection of random variables with
various distributions are said to exist. To ensure that the resulting theory is
not vacuous, a theorem is proven to show that there exists an $\Omega$ and a
measure $\mu$ on $\Omega$ for which a suitably rich collection of random
variables can be constructed.

While this pencil-and-paper approach could be used in formalization, it is
inconvenient in several ways.  First, while random variables are formally
functions on the sample space, in practice %
we often treat them as if
they were elements of their codomain.  For example if $X$ and $Y$ are two real-valued random variables, then one writes $X + Y$ to mean the random variable
$\lambda \omega.\, X(\omega) + Y(\omega)$. Similarly, if $f
: \mathbb{R} \to \mathbb{R}$, we write $f(X)$ for the random variable $(\lambda
\omega.\, f(X(\omega))$. While this kind of convention is well understood on paper,
trying to overload notations in a theorem prover to support it seems difficult.

The Giry monad~\citep{giry1982categorical} solves this problem by providing a syntactic sugar to describe
stochastic procedures concisely.

\subsection{Definition of the Giry Monad}

\newcommand{\mbind}{\texttt{bind}}
\newcommand{\mret}{\texttt{ret}}
\newcommand{\mretw}{\textsf{ret}}
\newcommand{\mdo}[3]{\texttt{do}\; #1\leftarrow#2\, ;\;#3}

The Giry monad is a triple $(\meas(.),\mbind,\mret)$.  For any measurable
space $X$, $\meas(X)$ is the space of probability measures over
$X$, that is, functions from measurable subsets of $X$ to $[0,1]$ that satisfy the additional axioms of probability measures. The function $\mbind$ is of type $\meas(X)
\rightarrow (X \rightarrow \meas(Y)) \rightarrow \meas(Y)$. That is,
it takes a probability measure on $X$, a function that transforms
values from $X$ into probability measures over $Y$, and returns a
probability measure on $Y$. The return function $\mret$
is of type $X \rightarrow \meas(X)$. It takes a value from
$X$ and returns a probability measure on $X$.
The \mathlib library defines this monad for general measures, which we
then restrict to probability measures.

Functions $\mathtt{bind}$ and $\mathtt{ret}$ construct probability
measures, so their definitions say what probability they assign to an
event. Letting $A$ be an event we define: %
\begin{align}
  \mbind\ \mu\ f\ A & = \int_{x \in X} f(x)(A) d\mu \\
  \mret \ x\ A & =
  \begin{cases}
    1 &\quad\text{if $x \in A$} \\
    0 &\quad\text{otherwise}
  \end{cases}
\end{align}
While we give the definitions here using standard mathematical notation, the formalization in \mathlib
uses the \lean definition of the integral. %
Here, $\mret(x)$ is the distribution that always returns $x$ with probability $1$.
The definition of $\mbind\ \mu\ f$ corresponds to first sampling 
from $\mu$ to obtain some value $x$, and then continuing with the probability measure $f(x)$.
Then, $\mbind$ and $\mret$ satisfy the usual monad laws 
\begin{align}
  \mbind\ (\mret\ x)\ f & = \mret(f\ x) \\
  \mbind\ \mu\ (\lambda x.\, \mret\ x) & = \mu\\
  \mbind\ (\mbind\ \mu\ f)\ g & = \mbind\ \mu\ (\lambda x. \mbind\ (f\ x)\ g)
\end{align}
However, these laws only hold when $f$ and $g$ are measurable functions. 
We will use the usual do-notation, writing $\mdo{x}{\mu}{g(x)}$ for $\mbind\ \mu\ g$.

As %
with any monad, we can define a function $\texttt{map}$ which lifts a function $f:
A \rightarrow B$, to a function of type
$\meas(A) \rightarrow \meas(B)$. Given $\mu: \meas(A)$, we interpret
$\texttt{map}\ f\ \mu$ as the probability distribution that first samples from
$\mu$ to obtain a value of type $A$, and then applies $f$ to it. Concretely, this
is defined in terms of $\mbind$ and $\mret$ as:
\begin{align}
\label{eqn:fmapdef}
\texttt{map}\ f\ \mu = \mdo{x}{\mu}{\mretw\ (f\ x)}
\end{align}

As an example, we show how to construct a distribution that samples $n$ independent
times from a distribution $\mu$ and returns the result as a tuple. That is,
we formally define the $\mu^n$ distribution that we %
used in \autoref{sec:decision-stumps}. Let
$(X,\mathcal{F})$ be a measurable space and $(X^n,\mathcal{F}^n)$ be
the measurable space where $X^n$ is the cartesian product of $X$ ($n$
times) and $\mathcal{F}^n$ is the product of $\mathcal{F}$ ($n$
times). Let $\mu$ be a probability measure on $(X,\mathcal{F})$. Then,
we define the measure $\mu^n$ recursively on $n$ as
\begin{align}
  \mu^1 &= \mu \\
  \mu^n &= \mdo{v}{\mu^{n-1}}{\mdo{s}{\mu}{\mretw(s,v)}}
\end{align}

Instead of nested tuples, as in the above, or lists of training examples,
as we did in \autoref{sec:decision-stumps}, we will use dependently typed vectors
in \lean. The following term (definition omitted) gives the probability measure corresponding to
taking \texttt{n+1} independent samples from a distribution and assembling them in
a vector:

\begin{verbatim}
def vec.prob_measure 
  (n : ℕ) (μ : probability_measure A) 
  : probability_measure (vec A n) := ...
\end{verbatim}

\subsection{Modeling Stump Training and Testing}

Let us now see how the decision stump training and testing procedures can be described using the Giry monad.
\renewcommand{\target}{{\tt target}}
\lean has a sectioning mechanism and a way to declare local variables. In
our formalization,
the lines below declare probability measures over the class of examples, and an arbitrary target threshold
value for labeling samples:
\begin{verbatim}
variables (μ: probability_measure ℍ) (target: ℍ) 
\end{verbatim}
When we write definitions that use these variables, \lean will interpret the definition to treat these variables as if they were additional parameters on which the function depends.\footnote{This is different from the behavior in Coq, where a definition that depends on section variables is only generalized to take additional arguments \emph{outside} the section where the variables are declared.}

The following pure function labels a sample according to the target.
\begin{verbatim}
def label (target: ℍ): ℍ → ℍ × bool :=
    λ x: ℍ, (x,rle x target)
\end{verbatim}
where {\texttt rle x target} returns true if $\texttt{x} \leq \texttt{\target}$ and false otherwise.

Finally, we can define the event that an example is mis-classified, and the $\error$ function: 
\begin{verbatim}
def error_set (h: ℍ) :=
  {x: ℍ | label h x ≠ label target x}

def error := λ h, μ (error_set h target)  
\end{verbatim}

The learning function $\alg$ will take a vector of labeled examples as input and output a hypothesis.
The function {\tt vec\_map} takes a function as
an argument and applies it pointwise to the elements of the vector. We use
this to label the inputs to the learning function:
\begin{verbatim}
def label_sample := vec_map (label target)
\end{verbatim}

Our learning algorithm starts by transforming any negative example to
0 and then stripping off the labels, with the following function:
\begin{verbatim}
def filter :=
  vec_map (λ p, if p.snd then p.fst else 0)
\end{verbatim}
This is safe since, if there were no positive examples,
the learning algorithm should return 0 as we described in \autoref{sec:decision-stumps}.
Finally, we can define the learning function $\alg$ that we will use. We call this function {\tt choose} in the formalization. It selects the largest example after the above filtering process:
\begin{verbatim}
def choose (n: ℕ):vec (ℍ × bool) n → ℍ :=
    λ data: (vec (ℍ × bool) n), max n (filter n data) 
\end{verbatim}

We then use the Giry monad to describe the measure on classifiers that results
from running the algorithm:
\begin{verbatim}
def denot: probability_measure ℍ :=
    let η := vec.prob_measure n μ  in 
    let ν := map (label_sample target n) η in  
    let γ := map (choose n) ν in
    γ
\end{verbatim}
Note that {\tt map} here is the monadic map operation defined in \autoref{eqn:fmapdef},
not {\tt vec\_map}. Thus, $\eta$ is a distribution on training examples which is then transformed to $\nu$, a distribution on labeled training examples. Then $\nu$ is
transformed into a distribution $\gamma$ on thresholds by lifting {\tt choose}. 

Finally, we have the following formal version of \autoref{thm:informal}:
\begin{verbatim}
theorem choose_PAC:
    ∀ ε: nnreal, ∀ δ: nnreal, ∀ n: ℕ, 
    ε > 0 → ε < 1 → δ > 0 → δ < 1 →
    n > (complexity ε δ) →
     (denot μ target n) {h: ℍ | error μ target h ≤ ε}
       ≥ 1 - δ
\end{verbatim}
where \texttt{complexity} is the following function:
\begin{verbatim}
def complexity (ε: ℝ) (δ: ℝ) : ℝ :=
  (log(δ) / log(1 - ε)) - (1: nat)
\end{verbatim}
The following sections outline how we prove this theorem in \lean.

\section{Deterministic Reasoning}
\label{sec:deterministic}
\label{sec:pure}

The overall proof builds on two simple assumptions that must be
satisfied by the learning algorithm. First, the algorithm must return
an estimate that is $\leq {\tt target}$.
Formally,
\begin{verbatim}
lemma choose_property_1: 
    ∀ n: ℕ, ∀ S: vec ℍ n,
    choose n (label_sample target n S) ≤ target
\end{verbatim}
Our implementation satisfies this property since after the ${\tt filter}$ step
in ${\tt choose}$, all of the examples will have been mapped to a value $\leq {\tt target}$, and we then
select the maximum value from the vector.

Second, the algorithm must return an estimate that is greater or equal
to any positive example. This is because the proof uses the assumption
that no examples lie in the region between the estimate and the
target. Formally,
\begin{verbatim}
lemma choose_property_2:
    ∀ n: ℕ, ∀ S: vec ℍ n, 
    ∀ i, 
    ∀ p = kth_projn (label_sample target n S) i,
    p.snd = tt → 
    p.fst ≤ choose n (label_sample target n S)
\end{verbatim}
where {\tt kth\_projn\ l\ i} is an expression giving component {\tt i} of the vector ${\tt l}$.
Our implementation satisfies this property because ${\tt choose}$ calls ${\tt filter}$ which leaves positive examples
unchanged, and then we select the maximum from the filtered vector.

These proofs are trivial and account for only a very small fraction of
the overall proof. Yet, these are the two specific properties of the algorithm that we need.

\section{Measurability Considerations}
\label{sec:measurability}

About a quarter of the formalization consists in proving that
various sets and functions are measurable. The predicate {\tt
is\_measurable S} states that the set {\tt S} is a measurable set,
while {\tt measurable f} states that the function {\tt f} is measurable.
These proofs can be long but are generally routine, with a few notable exceptions.

We will need to divide up the sample space into various intervals.
If $a$ and $b$ are two reals, then we write {\tt Ioo a b}, {\tt Ioc a b}, {\tt Ico a b}, {\tt Icc a b} in \lean to refer to the intervals $(a, b)$, $(a, b]$, $[a, b)$, and $[a, b]$, respectively.
To start, we observe that the error of a hypothesis is the measure of the interval
between it and the target.
\begin{verbatim}
lemma error_interval_1:
    ∀ h, h ≤ target →
    error μ target h = μ (Ioc h target)
lemma error_interval_2:
    ∀ h, h > target →
    error μ target h = μ (Ioc target h)
\end{verbatim}
We next use these lemmas to prove that 
the function that computes the $\error$ of a hypothesis is measurable:
\begin{verbatim}
lemma error_measurable:
    measurable (error μ target)
\end{verbatim}
\begin{proof}
  First, note that if $A$ and $B$ are measurable subsets such that $A \subseteq B$, then $\mu\, (B \setminus A) = \mu\, B - \mu\, A$.
  If $h \leq {\tt target}$, then
  \begin{align*}
  {\tt error\ μ\ target\ h} &= \mu\, (h, {\tt target}] \\
   &= \mu\, [0, {\tt target}] - \mu\, [0, h]
  \end{align*}
  Likewise, if $h > {\tt
    target}$ then {\tt error μ target h} = $\mu\, [0,h] - \mu\, [0,{\tt
      target}]$.
  Subtraction is measurable and testing whether a value is $\leq {\target}$ is measurable. Therefore, since measurability is closed under composition,
  it suffices to show that the function $\lambda
  x. \mu\, [0,x]$ is Borel measurable.
  Because this function is monotone, its measurability is a standard result, though this fact was missing from \mathlib.
\end{proof}

Next, one must show that the learning algorithm \texttt{choose} is measurable, after fixing the number of input examples:
\begin{verbatim}
lemma choose_measurable: measurable (choose n)
\end{verbatim}
\begin{proof}
  To prove that {\tt choose} is a measurable
  function, we must prove that {\tt max} is a measurable function.
  Because {\tt max} is continuous, it is Borel measurable.
\end{proof}

Although the previous proof is straightforward, it hinges on the fact that the
sigma-algebra structure we associate with \texttt{vec ℍ n} is the Borel
sigma-algebra. But, because we define a vector as an iterated product, another
possible sigma-algebra structure for \texttt{vec ℍ n} is the
\texttt{n+1}-ary product sigma-algebra.

Recall from the previous section that our development uses \lean's typeclass
mechanism to automatically associate product sigma-algebras with product
spaces, and Borel sigma-algebras with topological spaces. As the preceding paragraph explains, for \texttt{vec ℍ n}
there are two possible choices. Which choice should be used?
In programming languages with typeclasses, the problem of having to select
between two potentially different instances of a typeclass is called
a \emph{coherence} problem~\citep{peytonjones1997type}. Because of this ambiguity, \mathlib is careful
to only enable certain instances by default. Of course, this same potential ambiguity arises
in normal mathematical writing, when we omit mentioning the associated sigma-algebra.

Fortunately, in the case of \texttt{vec ℍ n},
these two sigma-algebras happen to be the same.  In general, if $X$ and $Y$ are
topological spaces with a \emph{countable basis}, then 
the Borel sigma-algebra on $X \times
Y$ is equal to the product of the Borel sigma-algebras on $X$ and $Y$.
The standard topology on the nonnegative reals has a countable basis, so
the equivalence holds.
Thus, although the proof of measurability for $\max$ can be simple, it uses
a subtle fact that resolves the ambiguity involved in referring to sets
without constantly mentioning their sigma-algebras.

\section{Probabilistic Reasoning}\label{section:walkthrough}
\label{sec:probabilistic}

The remainder of the proof involves the construction of the point $\theta$ and explicitly bounding the probability of various events.

Recall from the informal sketch in \autoref{sec:informal}
that we first case split on whether
$ \Pr_{x \sampled \mu}\left(x \leq \target\right) \leq \epsilon$ or not. In the language of measures,
this is equivalent to a case split on whether $\mu\, [0, \target] \leq \epsilon$. In the formalization,
it simplifies slightly the application of certain lemmas if we instead split on
whether $\mu\, (0, \target] \leq \epsilon$.
The following lemma is the key property in the case where the weight
between 0 and {\tt target} is $\leq \epsilon$. In that
case, the learning algorithm always chooses a hypothesis with error
at most $\epsilon$.
\begin{verbatim}
lemma always_succeed:
    ∀ ε: nnreal, ε > 0 → ∀ n: ℕ, 
    μ (Ioc 0 target) ≤ ε →  
    ∀ S: vec ℍ n, 
    error μ target 
            (choose n (label_sample target n S))
    ≤ ε
\end{verbatim}
\begin{proof}
By {\tt error\_interval\_1}, we know that the error is going to be equal to
the measure of the interval%
\begin{verbatim}
(Ioc (choose n (label_sample target n S)) target)
\end{verbatim}
Because we know {\tt choose} must return a threshold between $0$ and {\tt target}, this interval is a subset of {\tt (Ioc 0 target)}.
Since measures are monotone, this means the measure of that interval must be $\leq$ {\tt μ (Ioc 0 target)},
which is $\leq \epsilon$ by assumption.
\end{proof}

For the case where $\mu\, (0, \target] > \epsilon$, the  informal sketch
selected a point $\theta$ such that $\mu\, [\theta, {\tt \target}] = \epsilon$.
However, as we saw in \autoref{counterex:theta},
such a $\theta$ may not exist when $\mu$ is not continuous. Instead, we construct $\theta$ so that
$\mu\, [\theta, {\tt \target}] \geq \epsilon$, and 
$\mu\, (\theta, {\tt \target}] \leq \epsilon$. The following theorem
states the existence of such a point:
\begin{verbatim}
theorem extend_to_epsilon_1:
∀ ε: nnreal, ε > 0 → 
μ (Ioc 0 target) > ε →
∃ θ: nnreal, μ (Icc θ target) ≥ ε ∧
             μ (Ioc θ target) ≤ ε
\end{verbatim}
\begin{proof}
We take $\theta$ to be $\sup \{ x \in \support \mid \mu\, [x,\target] \geq \epsilon\}$. The supremum exists
because the set in question is bounded above by $\target$, and the set is nonempty because it must contain $0$ by our assumption that $\mu\, (0,\target] > \epsilon$.
To see that $\mu\, [\theta, \target] \geq \epsilon$, we can construct an increasing sequence of points $x_n \leq \theta$ such that $\lim_{n\to \infty} x_n = \theta$, where for each $n$, $\mu\, [x_n, \target] \geq \epsilon$. We have then that:
\[     \bigcap_{i} [x_i, \target] =  [\theta, \target] \]
We use this in conjunction with the fact that measures are continuous from above, meaning that if $A_1, A_2, \dots$ is a sequence of measurable
sets such that $A_{i+1} \subseteq A_{i}$ for all $i$, then
\[ \mu\, \left(\bigcap_{i=1}^{\infty} A_i\right) = \lim_{i \to \infty} \mu\, A_i \]
Hence we have
\begin{align*}
\mu\, [\theta, \target] &= \mu\, \left(\bigcap_{i=1}^{\infty} [x_i, \target]\right) \\
&= \lim_{n\to\infty} \mu\, [x_n, \target] \\
&\geq \epsilon
\end{align*}
The proof that $\mu\, (\theta, \target] \leq \epsilon$ is the dual argument, using continuity from below.
\end{proof}

The conclusion of this theorem states two inequalities involving $\theta$.  On
the one hand, we need $\theta$ to be small enough that we can ensure at least
one \emph{training example} will lie between $\theta$ and $\target$. On the other hand,
we want $\theta$ to be large enough that if we \emph{only} misclassify \emph{test 
examples} that lie between $\theta$ and $\target$, the error will
nevertheless be at most $\epsilon$.

The next two lemmas formalize these properties.
Recall that ${\tt choose}$ maps all negative training examples to $0$, leaves positive
examples unchanged, and then takes the maximum of the resulting vector.  The
next lemma says that given a point $\theta$ such that $\mu\, [\theta, {\tt
target}] \geq \epsilon$, the measure of the event that an example gets
mapped to something less than $\theta$ is at most $1 - \epsilon$.
\begin{verbatim}
lemma miss_prob:
    ∀ ε, ∀ θ: nnreal, θ > 0 → 
    μ (Icc θ target) ≥ ε →
    μ {x : ℍ | ∀ a b, 
         (a,b) = label target x → 
         (if b then a else 0) < θ} ≤ 1 - ε
\end{verbatim}

The next lemma shows why the property $\mu\, (\theta, \target] \leq \epsilon$ is
useful.  In particular, it says that for such a $\theta$, in order to have an
error $> \epsilon$ on the hypothesis selected by {\tt choose}, all training
examples must get mapped to something less than $\theta$. Formally, we say that the set of training
samples which would lead to an error greater than $\epsilon$, is a subset
of those in which all the examples get mapped to a value less than $\theta$. 
\begin{verbatim}
lemma all_missed: 
 ∀ ε: nnreal, 
 ∀ θ: nnreal, 
 μ (Ioc θ target) ≤ ε → 
 {S | error μ target
                  (choose n (label_sample target n S))
        > ε} ⊆ 
 {S |  ∀ i, 
       ∀ p = label target (kth_projn S i), 
       (if p.snd then p.fst else 0) < θ}
\end{verbatim}
Finally, we prove a bound related to the {\tt complexity} function, which computes the number
of training examples needed:
\begin{verbatim}
lemma complexity_enough:
    ∀ ε: nnreal, ∀ δ: nnreal, ∀ n: ℕ, 
    ε > (0: nnreal) → ε < (1: nnreal) → 
    δ > (0: nnreal) → δ < (1: nnreal) → 
    (n: ℝ) > (complexity ε δ) → ((1 - ε)^(n+1)) ≤ δ 
\end{verbatim}

Combining these lemmas together, we can finish the proof:

\begin{proof}[Proof of {\tt choose\_PAC}]

We have seen that {\tt always\_succeed} handles the case %
$\mu\, (0, {\tt target}] \leq \epsilon$.
For the other case, where
$\mu\, (0, {\tt target}] > \epsilon$, we can apply {\tt extend\_to\_epsilon\_1} to get a
$\theta$ with the specified properties.
By {\tt all\_missed} we know that the event that the hypothesis selected
has error $> \epsilon$ is a subset of the event where all the training
examples get mapped to $< \theta$. Then, by {\tt miss\_prob}
we know the probability that a given example gets mapped to $< \theta$ is 
$\leq 1 - \epsilon$. Because the training examples are selected independently, the probability
that all $n+1$ examples get mapped to a value $< \theta$ is at most
$(1-\epsilon)^{n+1}$.  Applying {\tt complexity\_enough}, we have that $(1
- \epsilon)^{n+1} \leq \delta$, so we are done. 
\end{proof}

\section{Related Work}\label{section:related-work}

Classic results about the average case behavior of quicksort and binary search
trees have been formalized by a number of authors using different proof
assistants~\citep{WeegenM08, EberlVT, TassarottiH17}. In each case, the
authors write down the algorithm to be analyzed using a variant of the monadic
style we discuss in \autoref{sec:giry}. \citet{GopinathanS20} verify the error rate of Bloom Filters and variants. \citet{AffeldtHS14} formalize results from information theory about lossy encoding.
For the most part, these formalizations
only use discrete probability theory, with the exception of Eberl et al.'s
analysis of treaps~\citep{EberlVT}, which requires general measure-theoretic probability.
They report that dealing with measurability issues adds some
overhead compared to pencil-and-paper reasoning, though they are able to
automate many of these proofs.

Several projects have formalized results from cryptography, which also involves
probabilistic reasoning~\citep{PetcherM15, BartheDGKSS13, BartheGB09, Blanchet06}. A
challenge in formalizing such proofs lies in the need to establish a relation
between the behavior of two different randomized algorithms, as part of the
game-playing approach to cryptographic security proofs.
Because cryptographic proofs generally only use
discrete probability theory, these libraries do not formalize measure-theoretic
results. There are many connections between cryptography and learning theory~\citep{Rivest91},
which would be interesting to formalize.

There have been formalizations of measure-theoretic probability
theory in a few proof assistants. \citet{hurd_thesis} formalized basic measure
theory in the HOL proof assistant, including a proof of Caratheodory's
extension theorem. \citet{HolzlH11} developed a more substantial library in the
Isabelle theorem prover, which has since been extended
further. \citet{AvigadHs17} used this library to formalize a proof of the
Central Limit Theorem.
Several measure theory libraries have also been developed in Coq~\citep{coq-proba, coq-markov, mathcomp-analysis}.
The ALEA library~\citep{AudebaudP09} instead uses a synthetic approach to discrete probability in Coq, a technique that has subsequently been extended to continuous probabilities by \citet{Bidlingmaier19}.

More recent work has formalized theoretical machine learning
results. \citet{selsam2017bugfree} use \lean to prove the correctness of an
optimization procedure for stochastic computation graphs. They prove that the
random gradients used in their stochastic backpropagation implementation are
unbiased. In their proof, they add axioms to the system for various basic mathematical facts. They argue that even if there are errors in these axioms that
could potentially lead to inconsistency, the process of constructing formal proofs
for the rest of the algorithm still helps eliminate mistakes.

\citet{bagnall2019certifying} use Coq to give machine-checked proofs of bounds on
generalization errors. They use Hoeffding's inequality to obtain bounds on error when the
hypothesis space is finite or there is a separate test-set on which to evaluate a classifier after training. They apply
this result to bound the generalization error of ReLU neural networks with
quantized weights. Their proof is restricted to discrete distributions and adds
some results from probability theory as axioms (Pinsker's inequality and Gibbs'
inequality).

\citet{bentkamp2019formal} use Isabelle/HOL to formalize a result by \citet{CohenSS16}, which shows
that deep convolutional arithmetic circuits are more expressive than
shallow ones, in the sense that shallow networks must be exponentially
larger in order to express the same function. Although convolutional
arithmetic circuits are not widely used in practice compared to other
artificial neural networks, this result is part of an effort to
understand theoretically the success of deep
learning. \citeauthor{bentkamp2019formal} report that they proved
a stronger version of the original result, and doing so allowed them
to structure the formal proof in a more modular way. The formalization
was completed only 14 months after the original arXiv posting
by \citeauthor{CohenSS16}, suggesting that once the right libraries
are available for a theorem prover, it is feasible to mechanize state
of the art results in some areas of theoretical machine learning in a
relatively brief period of time.

After the development described by %
our paper was publicly released,
\citet{mzinkevi} published a \lean library for probability
theory and theoretical machine learning. Among other results, this library contains
theorems about PAC learnability when the class of hypotheses is finite.
Because the decision stump hypothesis class is the set of all nonnegative
real numbers, our result is not covered by these theorems.

A related but distinct line of work applies machine learning techniques to
automatically construct formal proofs of theorems. Traditional approaches to
automated theorem proving rely on a mixture of heuristics and specialized
algorithms for decidable sub-problems. By using a pre-existing
corpus of formal proofs, supervised learning algorithms can be trained
to select hypotheses and construct proofs in a formal system~\citep{holist, gamepad, holstep, SelsamB19, KaliszykUMO18, JakubuvU19}.
 
\section{Conclusion}

We have presented a machine-checked, formal proof of PAC learnability
of the concept class of decision stumps. The proof is formalized using
the \lean theorem prover. We used the Giry monad to keep
the formalization simple and close to a pencil-and-paper proof. To
formalize this proof, we specialized the measure theory formalization
of the \mathlib library to the necessary basic probability theory. As
expected, the formalization is at times subtle when we must consider
topological or measurability results, mostly to prove that the
learning algorithm and $\error$ are measurable
functions. The most technical part of the proof
has to do with proving the existence of an interval
with the appropriate measure, a detail that %
standard textbook proofs either omit or get wrong.

Our work shows that the \lean prover and the
\mathlib library are mature enough to tackle a simple but classic
result in statistical learning theory. A next step would be to formally
prove more general results from VC-dimension theory.
In addition, there exist a number of
generalizations of PAC learnability, such as \emph{agnostic} PAC learnability, which removes the assumption that some hypothesis in the class perfectly classifies the examples. Other generalizations
allow for more than two classification labels and different kinds of $\error$
functions. It would be interesting to formalize these various extensions and some related applications.

\begin{acks}
We thank Gordon Stewart for comments on a previous draft of the paper. We thank the anonymous reviewers from the CPP’21 PC for their feedback. Some of the work described in this paper was performed while Koundinya Vajjha was an intern at Oracle Labs. Vajjha was additionally supported by the Alfred P. Sloan Foundation under grant number G-2018-10067. Banerjee’s research was based on work supported by the US National Science Foundation (NSF), while working at the Foundation. Any opinions, findings, and conclusions or recommendations expressed in the material are those of the authors and do not necessarily reflect the views of the NSF.
\end{acks}

\balance
\bibliographystyle{ACM-Reference-Format}
\bibliography{ref}

%%% -*-BibTeX-*-
%%% Do NOT edit. File created by BibTeX with style
%%% ACM-Reference-Format-Journals [18-Jan-2012].

\begin{thebibliography}{42}

%%% ====================================================================
%%% NOTE TO THE USER: you can override these defaults by providing
%%% customized versions of any of these macros before the \bibliography
%%% command.  Each of them MUST provide its own final punctuation,
%%% except for \shownote{}, \showDOI{}, and \showURL{}.  The latter two
%%% do not use final punctuation, in order to avoid confusing it with
%%% the Web address.
%%%
%%% To suppress output of a particular field, define its macro to expand
%%% to an empty string, or better, \unskip, like this:
%%%
%%% \newcommand{\showDOI}[1]{\unskip}   % LaTeX syntax
%%%
%%% \def \showDOI #1{\unskip}           % plain TeX syntax
%%%
%%% ====================================================================

\ifx \showCODEN    \undefined \def \showCODEN     #1{\unskip}     \fi
\ifx \showDOI      \undefined \def \showDOI       #1{#1}\fi
\ifx \showISBNx    \undefined \def \showISBNx     #1{\unskip}     \fi
\ifx \showISBNxiii \undefined \def \showISBNxiii  #1{\unskip}     \fi
\ifx \showISSN     \undefined \def \showISSN      #1{\unskip}     \fi
\ifx \showLCCN     \undefined \def \showLCCN      #1{\unskip}     \fi
\ifx \shownote     \undefined \def \shownote      #1{#1}          \fi
\ifx \showarticletitle \undefined \def \showarticletitle #1{#1}   \fi
\ifx \showURL      \undefined \def \showURL       {\relax}        \fi
% The following commands are used for tagged output and should be
% invisible to TeX
\providecommand\bibfield[2]{#2}
\providecommand\bibinfo[2]{#2}
\providecommand\natexlab[1]{#1}
\providecommand\showeprint[2][]{arXiv:#2}

\bibitem[\protect\citeauthoryear{Affeldt, Cohen, Kerjean, Mahboubi, Rouhling,
  Sakaguchi, and Strub}{Affeldt et~al\mbox{.}}{2020}]%
        {mathcomp-analysis}
\bibfield{author}{\bibinfo{person}{Reynald Affeldt}, \bibinfo{person}{Cyril
  Cohen}, \bibinfo{person}{Marie Kerjean}, \bibinfo{person}{Assia Mahboubi},
  \bibinfo{person}{Damien Rouhling}, \bibinfo{person}{Kazuhiko Sakaguchi},
  {and} \bibinfo{person}{Pierre-Yves Strub}.} \bibinfo{year}{2020}\natexlab{}.
\newblock \bibinfo{title}{{math-comp} Analysis Library}.
\newblock \bibinfo{howpublished}{\url{https://github.com/math-comp/analysis}}.
\newblock


\bibitem[\protect\citeauthoryear{Affeldt, Hagiwara, and
  S{\'{e}}nizergues}{Affeldt et~al\mbox{.}}{2014}]%
        {AffeldtHS14}
\bibfield{author}{\bibinfo{person}{Reynald Affeldt}, \bibinfo{person}{Manabu
  Hagiwara}, {and} \bibinfo{person}{Jonas S{\'{e}}nizergues}.}
  \bibinfo{year}{2014}\natexlab{}.
\newblock \showarticletitle{Formalization of Shannon's Theorems}.
\newblock \bibinfo{journal}{\emph{J. Autom. Reason.}} \bibinfo{volume}{53},
  \bibinfo{number}{1} (\bibinfo{year}{2014}), \bibinfo{pages}{63--103}.
\newblock


\bibitem[\protect\citeauthoryear{Audebaud and Paulin{-}Mohring}{Audebaud and
  Paulin{-}Mohring}{2009}]%
        {AudebaudP09}
\bibfield{author}{\bibinfo{person}{Philippe Audebaud} {and}
  \bibinfo{person}{Christine Paulin{-}Mohring}.}
  \bibinfo{year}{2009}\natexlab{}.
\newblock \showarticletitle{Proofs of randomized algorithms in {Coq}}.
\newblock \bibinfo{journal}{\emph{Sci. Comput. Program.}} \bibinfo{volume}{74},
  \bibinfo{number}{8} (\bibinfo{year}{2009}), \bibinfo{pages}{568--589}.
\newblock


\bibitem[\protect\citeauthoryear{Aumann}{Aumann}{1961}]%
        {Aumann}
\bibfield{author}{\bibinfo{person}{Robert~J. Aumann}.}
  \bibinfo{year}{1961}\natexlab{}.
\newblock \showarticletitle{Borel structures for function spaces}.
\newblock \bibinfo{journal}{\emph{Illinois J. Math.}} \bibinfo{volume}{5},
  \bibinfo{number}{4} (\bibinfo{date}{12} \bibinfo{year}{1961}),
  \bibinfo{pages}{614--630}.
\newblock


\bibitem[\protect\citeauthoryear{Avigad, H{\"{o}}lzl, and Serafin}{Avigad
  et~al\mbox{.}}{2017}]%
        {AvigadHs17}
\bibfield{author}{\bibinfo{person}{Jeremy Avigad}, \bibinfo{person}{Johannes
  H{\"{o}}lzl}, {and} \bibinfo{person}{Luke Serafin}.}
  \bibinfo{year}{2017}\natexlab{}.
\newblock \showarticletitle{A Formally Verified Proof of the Central Limit
  Theorem}.
\newblock \bibinfo{journal}{\emph{J. Autom. Reason.}} \bibinfo{volume}{59},
  \bibinfo{number}{4} (\bibinfo{year}{2017}), \bibinfo{pages}{389--423}.
\newblock


\bibitem[\protect\citeauthoryear{Bagnall and Stewart}{Bagnall and
  Stewart}{2019}]%
        {bagnall2019certifying}
\bibfield{author}{\bibinfo{person}{Alexander Bagnall} {and}
  \bibinfo{person}{Gordon Stewart}.} \bibinfo{year}{2019}\natexlab{}.
\newblock \showarticletitle{Certifying the True Error: Machine Learning in
  {Coq} with Verified Generalization Guarantees}. In
  \bibinfo{booktitle}{\emph{AAAI'19: The Thirty-Third AAAI Conference on
  Artificial Intelligence}}. \bibinfo{pages}{2662--2669}.
\newblock


\bibitem[\protect\citeauthoryear{Bansal, Loos, Rabe, Szegedy, and
  Wilcox}{Bansal et~al\mbox{.}}{2019}]%
        {holist}
\bibfield{author}{\bibinfo{person}{Kshitij Bansal}, \bibinfo{person}{Sarah~M.
  Loos}, \bibinfo{person}{Markus~N. Rabe}, \bibinfo{person}{Christian Szegedy},
  {and} \bibinfo{person}{Stewart Wilcox}.} \bibinfo{year}{2019}\natexlab{}.
\newblock \showarticletitle{{HOList}: An Environment for Machine Learning of
  Higher Order Logic Theorem Proving}. In
  \bibinfo{booktitle}{\emph{Thirty-sixth International Conference on Machine
  Learning (ICML)}}. \bibinfo{pages}{454--463}.
\newblock


\bibitem[\protect\citeauthoryear{Barthe, Dupressoir, Gr{\'{e}}goire, Kunz,
  Schmidt, and Strub}{Barthe et~al\mbox{.}}{2013}]%
        {BartheDGKSS13}
\bibfield{author}{\bibinfo{person}{Gilles Barthe},
  \bibinfo{person}{Fran{\c{c}}ois Dupressoir}, \bibinfo{person}{Benjamin
  Gr{\'{e}}goire}, \bibinfo{person}{C{\'{e}}sar Kunz},
  \bibinfo{person}{Benedikt Schmidt}, {and} \bibinfo{person}{Pierre{-}Yves
  Strub}.} \bibinfo{year}{2013}\natexlab{}.
\newblock \showarticletitle{EasyCrypt: {A} Tutorial}. In
  \bibinfo{booktitle}{\emph{Foundations of Security Analysis and Design {VII} -
  {FOSAD} 2012/2013 Tutorial Lectures}}. \bibinfo{pages}{146--166}.
\newblock


\bibitem[\protect\citeauthoryear{Barthe, Gr{\'{e}}goire, and
  B{\'{e}}guelin}{Barthe et~al\mbox{.}}{2009}]%
        {BartheGB09}
\bibfield{author}{\bibinfo{person}{Gilles Barthe}, \bibinfo{person}{Benjamin
  Gr{\'{e}}goire}, {and} \bibinfo{person}{Santiago~Zanella B{\'{e}}guelin}.}
  \bibinfo{year}{2009}\natexlab{}.
\newblock \showarticletitle{Formal certification of code-based cryptographic
  proofs}. In \bibinfo{booktitle}{\emph{POPL}}. \bibinfo{pages}{90--101}.
\newblock


\bibitem[\protect\citeauthoryear{Bentkamp, Blanchette, and Klakow}{Bentkamp
  et~al\mbox{.}}{2019}]%
        {bentkamp2019formal}
\bibfield{author}{\bibinfo{person}{Alexander Bentkamp},
  \bibinfo{person}{Jasmin~Christian Blanchette}, {and}
  \bibinfo{person}{Dietrich Klakow}.} \bibinfo{year}{2019}\natexlab{}.
\newblock \showarticletitle{A formal proof of the expressiveness of deep
  learning}.
\newblock \bibinfo{journal}{\emph{Journal of Automated Reasoning}}
  \bibinfo{volume}{63}, \bibinfo{number}{2} (\bibinfo{year}{2019}),
  \bibinfo{pages}{347--368}.
\newblock


\bibitem[\protect\citeauthoryear{Bidlingmaier, Faissole, and
  Spitters}{Bidlingmaier et~al\mbox{.}}{2019}]%
        {Bidlingmaier19}
\bibfield{author}{\bibinfo{person}{Martin~E. Bidlingmaier},
  \bibinfo{person}{Florian Faissole}, {and} \bibinfo{person}{Bas Spitters}.}
  \bibinfo{year}{2019}\natexlab{}.
\newblock \showarticletitle{Synthetic topology in Homotopy Type Theory for
  probabilistic programming}.
\newblock \bibinfo{journal}{\emph{CoRR}}  \bibinfo{volume}{abs/1912.07339}
  (\bibinfo{year}{2019}).
\newblock
\showeprint[arxiv]{1912.07339}
\urldef\tempurl%
\url{http://arxiv.org/abs/1912.07339}
\showURL{%
\tempurl}


\bibitem[\protect\citeauthoryear{Blanchet}{Blanchet}{2006}]%
        {Blanchet06}
\bibfield{author}{\bibinfo{person}{Bruno Blanchet}.}
  \bibinfo{year}{2006}\natexlab{}.
\newblock \showarticletitle{A Computationally Sound Mechanized Prover for
  Security Protocols}. In \bibinfo{booktitle}{\emph{2006 {IEEE} Symposium on
  Security and Privacy}}. \bibinfo{pages}{140--154}.
\newblock


\bibitem[\protect\citeauthoryear{Blumer, Ehrenfeucht, Haussler, and
  Warmuth}{Blumer et~al\mbox{.}}{1989}]%
        {blumer1989learnability}
\bibfield{author}{\bibinfo{person}{Anselm Blumer}, \bibinfo{person}{Andrzej
  Ehrenfeucht}, \bibinfo{person}{David Haussler}, {and}
  \bibinfo{person}{Manfred~K Warmuth}.} \bibinfo{year}{1989}\natexlab{}.
\newblock \showarticletitle{Learnability and the {Vapnik-Chervonenkis}
  dimension}.
\newblock \bibinfo{journal}{\emph{Journal of the ACM (JACM)}}
  \bibinfo{volume}{36}, \bibinfo{number}{4} (\bibinfo{year}{1989}),
  \bibinfo{pages}{929--965}.
\newblock


\bibitem[\protect\citeauthoryear{Cohen, Sharir, and Shashua}{Cohen
  et~al\mbox{.}}{2016}]%
        {CohenSS16}
\bibfield{author}{\bibinfo{person}{Nadav Cohen}, \bibinfo{person}{Or Sharir},
  {and} \bibinfo{person}{Amnon Shashua}.} \bibinfo{year}{2016}\natexlab{}.
\newblock \showarticletitle{On the Expressive Power of Deep Learning: {A}
  Tensor Analysis}. In \bibinfo{booktitle}{\emph{Proceedings of the 29th
  Conference on Learning Theory, {COLT} 2016}}. \bibinfo{pages}{698--728}.
\newblock


\bibitem[\protect\citeauthoryear{de~Moura, Kong, Avigad, van Doorn, and von
  Raumer}{de~Moura et~al\mbox{.}}{2015}]%
        {MouraKADR15}
\bibfield{author}{\bibinfo{person}{Leonardo~Mendon{\c{c}}a de Moura},
  \bibinfo{person}{Soonho Kong}, \bibinfo{person}{Jeremy Avigad},
  \bibinfo{person}{Floris van Doorn}, {and} \bibinfo{person}{Jakob von
  Raumer}.} \bibinfo{year}{2015}\natexlab{}.
\newblock \showarticletitle{The Lean Theorem Prover (System Description)}. In
  \bibinfo{booktitle}{\emph{{CADE-25} - 25th International Conference on
  Automated Deduction}}. \bibinfo{pages}{378--388}.
\newblock


\bibitem[\protect\citeauthoryear{Dudley}{Dudley}{2014}]%
        {dudley_uclt}
\bibfield{author}{\bibinfo{person}{R.~M. Dudley}.}
  \bibinfo{year}{2014}\natexlab{}.
\newblock \bibinfo{booktitle}{\emph{Uniform Central Limit Theorems}
  (\bibinfo{edition}{2nd} ed.)}.
\newblock \bibinfo{publisher}{Cambridge University Press}.
\newblock


\bibitem[\protect\citeauthoryear{Eberl, Haslbeck, and Nipkow}{Eberl
  et~al\mbox{.}}{2018}]%
        {EberlVT}
\bibfield{author}{\bibinfo{person}{Manuel Eberl}, \bibinfo{person}{Max~W.
  Haslbeck}, {and} \bibinfo{person}{Tobias Nipkow}.}
  \bibinfo{year}{2018}\natexlab{}.
\newblock \showarticletitle{Verified Analysis of Random Trees}. In
  \bibinfo{booktitle}{\emph{ITP}}. \bibinfo{pages}{196--214}.
\newblock


\bibitem[\protect\citeauthoryear{Giry}{Giry}{1982}]%
        {giry1982categorical}
\bibfield{author}{\bibinfo{person}{Mich\`{e}le Giry}.}
  \bibinfo{year}{1982}\natexlab{}.
\newblock \showarticletitle{A Categorical Approach to Probability Theory}. In
  \bibinfo{booktitle}{\emph{Categorical Aspects of Topology and Analysis}}
  \emph{(\bibinfo{series}{Lecture Notes in Mathematics},
  Vol.~\bibinfo{volume}{915})},
  \bibfield{editor}{\bibinfo{person}{B.~Banaschewski}} (Ed.).
  \bibinfo{pages}{68--85}.
\newblock


\bibitem[\protect\citeauthoryear{Gopinathan and Sergey}{Gopinathan and
  Sergey}{2020}]%
        {GopinathanS20}
\bibfield{author}{\bibinfo{person}{Kiran Gopinathan} {and}
  \bibinfo{person}{Ilya Sergey}.} \bibinfo{year}{2020}\natexlab{}.
\newblock \showarticletitle{Certifying Certainty and Uncertainty in Approximate
  Membership Query Structures}. In \bibinfo{booktitle}{\emph{CAV}},
  \bibfield{editor}{\bibinfo{person}{Shuvendu~K. Lahiri} {and}
  \bibinfo{person}{Chao Wang}} (Eds.). \bibinfo{pages}{279--303}.
\newblock


\bibitem[\protect\citeauthoryear{H{\"{o}}lzl}{H{\"{o}}lzl}{2013}]%
        {Holzl13}
\bibfield{author}{\bibinfo{person}{Johannes H{\"{o}}lzl}.}
  \bibinfo{year}{2013}\natexlab{}.
\newblock \emph{\bibinfo{title}{Construction and stochastic applications of
  measure spaces in higher-order logic}}.
\newblock \bibinfo{thesistype}{Ph.D. Dissertation}. \bibinfo{school}{Technical
  University Munich}.
\newblock


\bibitem[\protect\citeauthoryear{H{\"{o}}lzl and Heller}{H{\"{o}}lzl and
  Heller}{2011}]%
        {HolzlH11}
\bibfield{author}{\bibinfo{person}{Johannes H{\"{o}}lzl} {and}
  \bibinfo{person}{Armin Heller}.} \bibinfo{year}{2011}\natexlab{}.
\newblock \showarticletitle{Three Chapters of Measure Theory in
  {Isabelle/HOL}}. In \bibinfo{booktitle}{\emph{ITP}}.
  \bibinfo{pages}{135--151}.
\newblock


\bibitem[\protect\citeauthoryear{Huang, Dhariwal, Song, and Sutskever}{Huang
  et~al\mbox{.}}{2019}]%
        {gamepad}
\bibfield{author}{\bibinfo{person}{Daniel Huang}, \bibinfo{person}{Prafulla
  Dhariwal}, \bibinfo{person}{Dawn Song}, {and} \bibinfo{person}{Ilya
  Sutskever}.} \bibinfo{year}{2019}\natexlab{}.
\newblock \showarticletitle{GamePad: {A} Learning Environment for Theorem
  Proving}. In \bibinfo{booktitle}{\emph{7th International Conference on
  Learning Representations, {ICLR} 2019}}.
\newblock


\bibitem[\protect\citeauthoryear{Hurd}{Hurd}{2003}]%
        {hurd_thesis}
\bibfield{author}{\bibinfo{person}{Joe Hurd}.} \bibinfo{year}{2003}\natexlab{}.
\newblock \emph{\bibinfo{title}{Formal Verification of Probabilistic
  Algorithms}}.
\newblock \bibinfo{thesistype}{Ph.D. Dissertation}. \bibinfo{school}{Cambridge
  University}.
\newblock


\bibitem[\protect\citeauthoryear{Jakubuv and Urban}{Jakubuv and Urban}{2019}]%
        {JakubuvU19}
\bibfield{author}{\bibinfo{person}{Jan Jakubuv} {and} \bibinfo{person}{Josef
  Urban}.} \bibinfo{year}{2019}\natexlab{}.
\newblock \showarticletitle{Hammering Mizar by Learning Clause Guidance (Short
  Paper)}. In \bibinfo{booktitle}{\emph{ITP}}. \bibinfo{pages}{34:1--34:8}.
\newblock


\bibitem[\protect\citeauthoryear{Kaliszyk, Chollet, and Szegedy}{Kaliszyk
  et~al\mbox{.}}{2017}]%
        {holstep}
\bibfield{author}{\bibinfo{person}{Cezary Kaliszyk},
  \bibinfo{person}{Fran{\c{c}}ois Chollet}, {and} \bibinfo{person}{Christian
  Szegedy}.} \bibinfo{year}{2017}\natexlab{}.
\newblock \showarticletitle{{HolStep}: {A} Machine Learning Dataset for
  Higher-order Logic Theorem Proving}. In \bibinfo{booktitle}{\emph{5th
  International Conference on Learning Representations, {ICLR} 2017}}.
\newblock


\bibitem[\protect\citeauthoryear{Kaliszyk, Urban, Michalewski, and
  Ols{\'{a}}k}{Kaliszyk et~al\mbox{.}}{2018}]%
        {KaliszykUMO18}
\bibfield{author}{\bibinfo{person}{Cezary Kaliszyk}, \bibinfo{person}{Josef
  Urban}, \bibinfo{person}{Henryk Michalewski}, {and} \bibinfo{person}{Miroslav
  Ols{\'{a}}k}.} \bibinfo{year}{2018}\natexlab{}.
\newblock \showarticletitle{Reinforcement Learning of Theorem Proving}. In
  \bibinfo{booktitle}{\emph{NeurIPS}}. \bibinfo{pages}{8836--8847}.
\newblock


\bibitem[\protect\citeauthoryear{Kam}{Kam}{2008}]%
        {coq-markov}
\bibfield{author}{\bibinfo{person}{Robert Kam}.}
  \bibinfo{year}{2008}\natexlab{}.
\newblock \bibinfo{title}{{coq-markov} Library}.
\newblock \bibinfo{howpublished}{\url{https://github.com/coq-contribs/markov}}.
\newblock


\bibitem[\protect\citeauthoryear{Kearns and Vazirani}{Kearns and
  Vazirani}{1994}]%
        {kearns1994introduction}
\bibfield{author}{\bibinfo{person}{Michael~J Kearns} {and}
  \bibinfo{person}{Umesh~Virkumar Vazirani}.} \bibinfo{year}{1994}\natexlab{}.
\newblock \bibinfo{booktitle}{\emph{An {I}ntroduction to {C}omputational
  {L}earning {T}heory}}.
\newblock \bibinfo{publisher}{MIT press}.
\newblock


\bibitem[\protect\citeauthoryear{Mohri, Rostamizadeh, and Talwalkar}{Mohri
  et~al\mbox{.}}{2018}]%
        {mohri2018foundations}
\bibfield{author}{\bibinfo{person}{Mehryar Mohri}, \bibinfo{person}{Afshin
  Rostamizadeh}, {and} \bibinfo{person}{Ameet Talwalkar}.}
  \bibinfo{year}{2018}\natexlab{}.
\newblock \bibinfo{booktitle}{\emph{Foundations of {M}achine {L}earning}}.
\newblock \bibinfo{publisher}{MIT press}.
\newblock


\bibitem[\protect\citeauthoryear{Petcher and Morrisett}{Petcher and
  Morrisett}{2015}]%
        {PetcherM15}
\bibfield{author}{\bibinfo{person}{Adam Petcher} {and} \bibinfo{person}{Greg
  Morrisett}.} \bibinfo{year}{2015}\natexlab{}.
\newblock \showarticletitle{The Foundational Cryptography Framework}. In
  \bibinfo{booktitle}{\emph{POST}}. \bibinfo{pages}{53--72}.
\newblock


\bibitem[\protect\citeauthoryear{Peyton~Jones, Jones, and Meijer}{Peyton~Jones
  et~al\mbox{.}}{1997}]%
        {peytonjones1997type}
\bibfield{author}{\bibinfo{person}{Simon Peyton~Jones}, \bibinfo{person}{Mark
  Jones}, {and} \bibinfo{person}{Erik Meijer}.}
  \bibinfo{year}{1997}\natexlab{}.
\newblock \showarticletitle{Type classes: an exploration of the design space}.
  In \bibinfo{booktitle}{\emph{Haskell Workshop}}.
\newblock


\bibitem[\protect\citeauthoryear{Rivest}{Rivest}{1991}]%
        {Rivest91}
\bibfield{author}{\bibinfo{person}{Ronald~L. Rivest}.}
  \bibinfo{year}{1991}\natexlab{}.
\newblock \showarticletitle{Cryptography and Machine Learning}. In
  \bibinfo{booktitle}{\emph{Advances in Cryptology - {ASIACRYPT} '91}}.
  \bibinfo{pages}{427--439}.
\newblock


\bibitem[\protect\citeauthoryear{Selsam and Bj{\o}rner}{Selsam and
  Bj{\o}rner}{2019}]%
        {SelsamB19}
\bibfield{author}{\bibinfo{person}{Daniel Selsam} {and}
  \bibinfo{person}{Nikolaj Bj{\o}rner}.} \bibinfo{year}{2019}\natexlab{}.
\newblock \showarticletitle{Guiding High-Performance {SAT} Solvers with
  Unsat-Core Predictions}. In \bibinfo{booktitle}{\emph{Theory and Applications
  of Satisfiability Testing - {SAT} 2019}}. \bibinfo{pages}{336--353}.
\newblock


\bibitem[\protect\citeauthoryear{Selsam, Liang, and Dill}{Selsam
  et~al\mbox{.}}{2017}]%
        {selsam2017bugfree}
\bibfield{author}{\bibinfo{person}{Daniel Selsam}, \bibinfo{person}{Percy
  Liang}, {and} \bibinfo{person}{David Dill}.} \bibinfo{year}{2017}\natexlab{}.
\newblock \showarticletitle{Developing Bug-Free Machine Learning Systems With
  Formal Mathematics}. In \bibinfo{booktitle}{\emph{International Conference on
  Machine Learning (ICML)}}. \bibinfo{pages}{3047--3056}.
\newblock


\bibitem[\protect\citeauthoryear{Shalev-Shwartz and Ben-David}{Shalev-Shwartz
  and Ben-David}{2014}]%
        {shalev2014understanding}
\bibfield{author}{\bibinfo{person}{Shai Shalev-Shwartz} {and}
  \bibinfo{person}{Shai Ben-David}.} \bibinfo{year}{2014}\natexlab{}.
\newblock \bibinfo{booktitle}{\emph{Understanding {M}achine {L}earning: From
  {T}heory to {A}lgorithms}}.
\newblock \bibinfo{publisher}{Cambridge University Press}.
\newblock


\bibitem[\protect\citeauthoryear{Tassarotti}{Tassarotti}{2020}]%
        {coq-proba}
\bibfield{author}{\bibinfo{person}{Joseph Tassarotti}.}
  \bibinfo{year}{2020}\natexlab{}.
\newblock \bibinfo{title}{{coq-proba} Probability Library}.
\newblock
  \bibinfo{howpublished}{\url{https://github.com/jtassarotti/coq-proba}}.
\newblock


\bibitem[\protect\citeauthoryear{Tassarotti and Harper}{Tassarotti and
  Harper}{2018}]%
        {TassarottiH17}
\bibfield{author}{\bibinfo{person}{Joseph Tassarotti} {and}
  \bibinfo{person}{Robert Harper}.} \bibinfo{year}{2018}\natexlab{}.
\newblock \showarticletitle{Verified Tail Bounds for Randomized Programs}. In
  \bibinfo{booktitle}{\emph{ITP}}. \bibinfo{pages}{560--578}.
\newblock


\bibitem[\protect\citeauthoryear{{The mathlib Community}}{{The mathlib
  Community}}{2020}]%
        {mathlib-cpp}
\bibfield{author}{\bibinfo{person}{{The mathlib Community}}.}
  \bibinfo{year}{2020}\natexlab{}.
\newblock \showarticletitle{The {L}ean {M}athematical {L}ibrary}. In
  \bibinfo{booktitle}{\emph{CPP}}. \bibinfo{pages}{367--381}.
\newblock


\bibitem[\protect\citeauthoryear{Valiant}{Valiant}{1984}]%
        {Valiant84}
\bibfield{author}{\bibinfo{person}{Leslie~G. Valiant}.}
  \bibinfo{year}{1984}\natexlab{}.
\newblock \showarticletitle{A Theory of the Learnable}.
\newblock \bibinfo{journal}{\emph{Commun. {ACM}}} \bibinfo{volume}{27},
  \bibinfo{number}{11} (\bibinfo{year}{1984}), \bibinfo{pages}{1134--1142}.
\newblock


\bibitem[\protect\citeauthoryear{van~der Weegen and McKinna}{van~der Weegen and
  McKinna}{2008}]%
        {WeegenM08}
\bibfield{author}{\bibinfo{person}{Eelis van~der Weegen} {and}
  \bibinfo{person}{James McKinna}.} \bibinfo{year}{2008}\natexlab{}.
\newblock \showarticletitle{A Machine-Checked Proof of the Average-Case
  Complexity of Quicksort in {Coq}}. In \bibinfo{booktitle}{\emph{TYPES}}.
  \bibinfo{pages}{256--271}.
\newblock


\bibitem[\protect\citeauthoryear{Vapnik}{Vapnik}{2000}]%
        {vapnik2000}
\bibfield{author}{\bibinfo{person}{Vladimir~Naumovich Vapnik}.}
  \bibinfo{year}{2000}\natexlab{}.
\newblock \bibinfo{booktitle}{\emph{The Nature of Statistical Learning Theory,
  Second Edition}}.
\newblock \bibinfo{publisher}{Springer}.
\newblock
\showISBNx{978-0-387-98780-4}


\bibitem[\protect\citeauthoryear{Zinkevich}{Zinkevich}{2020}]%
        {mzinkevi}
\bibfield{author}{\bibinfo{person}{Martin Zinkevich}.}
  \bibinfo{year}{2020}\natexlab{}.
\newblock
\newblock
\urldef\tempurl%
\url{https://github.com/google/formal-ml}
\showURL{%
\tempurl}


\end{thebibliography}

\end{document}